\documentclass[graybox]{svmult}

\usepackage{type1cm}        
\usepackage{makeidx}         
\usepackage{graphicx}        
\usepackage{multicol}        
\usepackage[bottom]{footmisc}

\usepackage{newtxtext}       %
\usepackage{newtxmath}       
\usepackage{algorithm}			
\usepackage[noend]{algpseudocode}
\usepackage{caption}
\usepackage[most]{tcolorbox}
\usepackage{booktabs}
\usepackage{makecell}
\usepackage{mdframed}
\usepackage{xcolor}
\usepackage{multirow}
\usepackage{siunitx}
\usepackage{longtable}
\usepackage{newlfont}

\makeindex             

\begin{document}

\title*{Route Planning Using Nature Inspired Algorithms}
\author{Priyansh Saxena, Raahat Gupta and Akshat Maheshwari}
\institute{Priyansh Saxena \at ABV-Indian Institute of Information Technology and Management, Gwalior; \email{saxenapriyanshasd@gmail.com},
\and Raahat Gupta \at ABV-Indian Institute of Information Technology and Management, Gwalior; \email{raahat.gupta.1998@gmail.com},
\and Akshat Maheshwari \at ABV-Indian Institute of Information Technology and Management, Gwalior; \email{aks3d76@gmail.com}}
%
%
\maketitle

\abstract*{There are many different heuristic algorithms for solving combinatorial optimization problems that are commonly described as Nature-Inspired Algorithms (NIAs). Generally, they are inspired by some natural phenomenon, and due to their inherent converging and stochastic nature, they are known to give optimal results when compared to classical approaches. There are a large number of applications of NIAs, perhaps the most popular being route planning problems in robotics – problems that require a sequence of translation and rotation steps from start to the goal in an optimized manner while avoiding obstacles in the environment. In this chapter, we will first give an overview of Nature-Inspired Algorithms followed by their classification and common examples. We will then discuss how the NIAs have applied to solve the route planning problem.}

\abstract{There are many different heuristic algorithms for solving combinatorial optimization problems that are commonly described as Nature-Inspired Algorithms (NIAs). Generally, they are inspired by some natural phenomenon, and due to their inherent converging and stochastic nature, they are known to give optimal results when compared to classical approaches. There are a large number of applications of NIAs, perhaps the most popular being route planning problems in robotics – problems that require a sequence of translation and rotation steps from start to the goal in an optimized manner while avoiding obstacles in the environment. In this chapter, we will first give an overview of Nature-Inspired Algorithms followed by their classification and common examples. We will then discuss how the NIAs have applied to solve the route planning problem.}

\section{Introduction}
\label{sec:1}

Man has always looked towards nature as an origin of insight to answer the most complex of questions that surround us. Many complex physical phenomena that first seem mystifying or baffling at least, can be found already being produced and controlled in nature. A careful study of these can lead to the development of algorithms and theories can be used to solve complex problems in our macro world. The techniques that are inspired either directly or indirectly from processes observed in nature are called \textbf{Nature-Inspired Algorithms}, and their applications in effective route planning will be the main focus of this chapter.

Every natural organism living today has gone through centuries and millennia of evolution. This has lead to the development of those body organs which are essential in the survival of the species in the long run, for example, turtles developed shells to protect against other creatures, birds evolved with wings to fly, and apes and early humans started using their front limbs to hold things. This several hundreds of millions of years of evolution are the primary reason we look for inspiration in the first place - for instance, we can solve complex transportation problems in toady's world by taking a closer look at how ants and other species look for their food in the wild.

This chapter takes a more in-depth look at various Nature-Inspired Algorithms(NIAs) and how we can apply them in optimisation problems, specifically route planning. We will also discuss the walk-through of a basic NIA, along with giving examples of several NIAs used in today's world. This chapter concludes by giving some insights into the motivation of taking inspirations from nature.


\section{The Problem of Route Planning}
\label{sec:2}

Humans have reached a great deal of automation in the twenty-first century. Examples of programmed robots range from the ones operating forklifts in Amazon warehouses, to aerial drones supplementing the delivery of medical supplies in Rwanda and Kenya, and rovers on the surface of Mars, Moon and other celestial bodies. All of these robots work in complex terrains, and they have to be programmed to automatically detect obstacles in their route, and take measures to avoid them.

The gist of the \textbf{route planning} problem is as follows - the agent/robot has to move from a starting location to its goal location while avoiding any/all obstacles in its route. Since obstacles in real-world can be, and often are, random and arbitrary, their positions cannot be accurately defined beforehand. Thus the robot has to \textit{search} for obstacles and develop alternate routes to avoid them on-the-fly. This is what makes the problem of route planning so tricky (it is actually NP-hard), and its immense applications in today's world mean that route planning has become a hot research topic in recent years.

A formal definition of route planning is given in the following section.

\subsection{Overview of Route Planning}
\label{subsec:2.1}

Route planning is considered as the process of identifying a sequence of transnational and rotational steps that the robot must take to reach destination location from a given source location in the shortest possible time and by avoiding collisions within the environment which contains static as well as moving obstacles. This problem becomes exceedingly challenging when the environment is made of dynamic obstacles because it requires the route to be re-planned in real-time when a new obstacle in discovered. Additional complexities include that the route should be simple, smooth and should be realistic to the degree that it can be followed by an unmanned aerial vehicle (UAV) in a practical scenario. Researchers generally consider two types of robots for route planning: \textit{land robots} are generally restricted to two-dimensional environments. On the other hand, \textit{air \& water robots} have more degrees of freedom and can change their position in any direction in space. Due to their complexity and immense practical applications, aerial robots have been studied the most.

Route-planning approaches can be classified on the basis of two approaches: $(i)$ \textbf{global route planning} also called \textit{offline approach}, and $(ii)$ \textbf{local route planning} or \textit{online approach} \cite{survey-1}. In global route planning, we create a high-level route based on a map whose current and past perceptive are known to the algorithm. It generally produces a low-dimensional and optimised route; however, it fails to react to unknown
obstacles. Online route planning algorithms do not input any previous information based on the environment. The route it gives, which is high-dimensional and low-level, is only over a fragment of the total environment. This fragment can be re-routed when a new obstacle arrives without altering the entire route. Hence, local planners are more suited for highly changing environments.

\subsection{Methods to Solve Route Planning Problems}
\label{subsec:2.2}

Several methods have been proposed to provide solutions for route planning problems. These are classified into: \textit{classical methods} and \textit{heuristic methods}. 

\subsubsection{Classical Methods}
\label{subsubsec:2.2.1}

Many classical methods for route planning have been proposed over the decades. Here, we will discuss four such methods: Cell Decomposition method (CD), Potential Field Method (PFM), Sub-goal method (SG) and Sampling-based methods,   .

\runinhead{Cell Decomposition Method} This method is based on the process of dividing the available free space in a robot's configuration into smaller regions, also known as \textit{cells}. This process gives a simple route collision-free route \cite{survey-2}. 

\runinhead{Potential Field Method} This method takes inspirations from the concept of potential field in electrostatics. The agent is analogous to a charged particle, and the goal is assigned an attractive charge. Thus, the agent will be attracted toward the goal. The obstacles, on the other hand, are assigned a negative force with respect to the agent. Thus, the agent will be repelled from the obstacles and move toward the goal \cite{survey-3}. The general formula for finding repelling potential is altered to give a reduced number of oscillations and avoiding improving practical concerns. An advantage of this method is that it can be successfully extended to multiple cruise UAVs as well.

\runinhead{Sub-goal Method} This method makes use of different attainable configurations from start to the goal, while bypassing all obstacles. This method is immensely popular for applications in robot navigation, as seen in \cite{survey-4}.

\runinhead{Sampling-Based motion Planning} These schemes are especially useful in complex real-life planning scenarios.
The most influential sampling-based methods consist of \textit{Rapidly-exploring Random Trees (RRT)} and \textit{Probabilistic Road-Map (PRM)} \cite{survey-5}. The similarity between these two methods is the concept of randomly-sampled connecting points, however they are different in the aspect that the graph which connects the points, is created.

Constant efforts are being made for applying classical methods to solve practical and real-world problems. Some of the algorithms that have shown promise include \textit{Configuration Space Collision Maps} and \textit{Voronoi Diagrams}. The working of these algorithms is left to the reader as a further study \cite{survey-6} \cite{survey-7} \cite{survey-8}. Disadvantages of classical approaches include their non-ability to produce optimal routes and tendency to get stuck in  local minima. The existence of multiple obstacles provides a hindrance to well-functioning of some of the environments. Hence, heuristic approaches are more prevalent today.

\begin{figure}[t]
\centering
\includegraphics[scale=0.18]{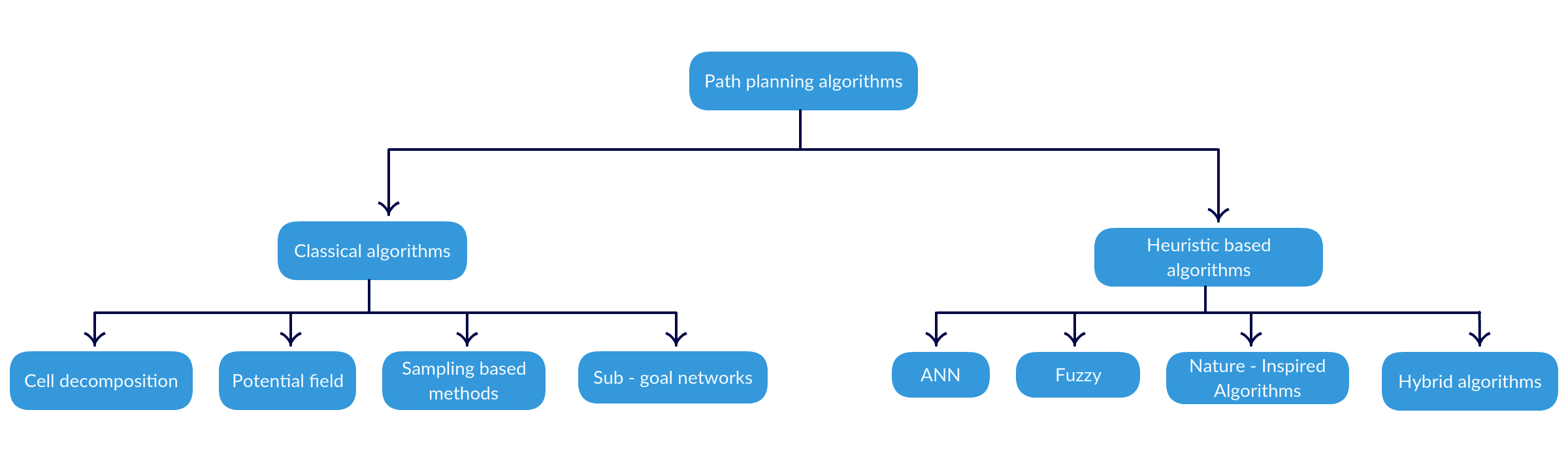}
\caption{The classification of robot route planning algorithms}
\label{fig:1}      
\end{figure}

\subsubsection{Heuristic Methods}
\label{subsubsec:2.2.2}

Heuristic methods are immensely popular in the field of route-planning research toady. Some of the standard methods are discussed below.

\runinhead{Neural Networks} The various algorithms that come under the umbrella of neural networks have seen an explosion in popularity in the past few years. These methods have also been applied to solve route planning problems \cite{survey-9}. Neural networks are known for finding out the relations between inputs and outputs. Their working, especially in robot navigation, is categorised as: $(i)$ interpreting input data, $(ii)$ enforcing techniques of obstacle avoidance $(iii)$ implementing route planning. A common neural network paradigm used for training \textbf{Recurrent Neural Networks (RNN)} is \textit{Reservoir Computing}. The model is made up of two randomly generated RNNs - one of which models the localisation ability and the other "learns" the navigation skill. A collision-free route can also be constructed using a combination of two neural network concepts. These are \textit{principal component analysis (PCA)} and a \textit{multi-layer perceptron (MLP)}.

\runinhead{Fuzzy Logic} The robot takes decisions with reference to a set of IF-THEN rules. It starts by diving the task into simpler problems \cite{survey-10}. The cost function is calculated and goals such as orientation of target, avoidance of obstacles and rotational movement are included in the above-mentioned cost function in order to determine the optimum steering angle $\theta$. The mobile robot varies the weights of the cost function and navigates through the environment.

\runinhead{Hybrid Algorithms} Researchers have integrated neural networks with their fuzzy counterparts in an effort to extract the positives of both the algorithms. This hybrid is also known to show positive results.

\runinhead{Nature Inspired Algorithms} NIAs are the most recent addition to this vast list of algorithms. These are generally more complex than the above-discussed algorithms and required more time and computation power. However, NIAs are shown to give more optimised and accurate answers than the above algorithms. As a result, NIAs will constitute a major portion of the remainder of this chapter. Most NIAs include the concept of \textbf{Swarm Intelligence (SI)}, which is the property of a system where the composite demeanour of unsophisticated agents interacting locally with their environment cause orderly functional global patterns to emerge \cite{survey-11}. Examples of NIAs may include Particle Swarm Optimisation (PSO), Salp Swarm Algorithm (SSA), Artificial Bee Colony Optimisation(ABC), and Ant Colony Optimisation(ACO) among others. These algorithms, along with the concept of Swarm Intelligence, will be discussed in further depths in the upcoming sections.


\section{Nature Inspired Algorithms}
\label{sec:3}

The scientific field of bionics helps us link biological processes, functions and organisational principles to modern technologies. There is a diverse range of algorithms, mathematical and meta-heuristic, developed specifically for the process of assigning expertise from the biological lifeforms to human life through technology. As a result, different kinds of optimisation algorithms have been developed which apply the concepts found in nature to practical and research life \cite{survey-12}. The techniques - Nature-Inspired Algorithms - will be the focus of this section.

\subsection{Objectives of NIA}
\label{subsec:3.1}

NIAs are designed with an objective of optimisation - i.e. of a problem of which multiple solutions exist, and the interest is to determine the solution with minimum cost. Usually, the cost here is the lowest time taken, but other factors such as lowest distance, or a combination of the two can be considered.

Here, we describe two key factors terms useful to understand the constraints of most optimisation problems: \textbf{exploration} and \textbf{exploitation}. Exploration consists of finding the global optima by searching the entire search space; as the name suggests, we are \textit{exploring} our surroundings in hopes of finding resources that are better than what we currently have. Exploitation, on the other hand, consists of finding the local optima in the explored solution space. We are \textit{exploiting} what all resources we have to get the best possible output. It is important to understand that these two processes are often applied simultaneously - intense exploration does not give optimal solution while deep exploitation traps an algorithm in local optima. Hence, a balance between these two processes is necessary \cite{survey-13}.

Nature-Inspired Algorithms are broadly classified into two categories: \textit{Genetic Algorithms (GA)} and \textit{Swarm Intelligence (SI).} These two categories are described in the next sections.

\subsection{Genetic Algorithms}
\label{subsec:3.2}

Genetic Algorithm (GA) was developed as a tool to understand the natural processes behind millions of years of evolution. It involves a study of processes like natural selection, crossover, and mutation - hence, they are also known as \textit{evolutionary algorithms}. Later these concepts are applied to optimisation problems and have their applications in route planning.

The major inspiration behind GA has been the theory "Survival of the Fittest" given by Charles Darwin \cite{survey-14}. These algorithms are able to apply natural selection, recombination and mutation so as to closely resemble the processes by living organisms, while at the same time giving impressive results in modern research. 

\subsection{Swarm Intelligence}
\label{subsec:3.3}

Swarm Intelligence (SI) mainly composes of three principles: \textit{evaluation}, \textit{comparing} and \textit{imitation}. The term evaluation denotes the capability to analyse the positive and negative effects of property in nature. Comparing comes naturally to living beings, wherein they compare themselves with other beings. The main purpose of these comparisons is to bring a sense of motivation to learn and/or modification. Imitation is defined as an effective form of learning.

\runinhead{Properties of a Swarm Intelligence System}
Following are the properties generally seen in a Swarm Intelligence (SI) system: 
\begin{enumerate}
\item A swarm is generally composed of many individual entities. Hence, the effects of outliers are diminished.
\item The individuals that make up the swarm are generally homogeneous - they are either identical or belong to only a few classifications.
\item Individuals interact among each other and share information about themselves or what they have learnt about the environment; this sharing is either direct or indirect.
\item These interactions between the individual "particles" constitute the overall behaviour of the system - the group behaviour self-organises.
\end{enumerate}

\runinhead{Principles Involved in Swarm Intelligence}
Swarm Intelligence can be described by taking into consideration the following principles \cite{survey-15}:
\begin{enumerate}
\item \textbf{Proximity Principle}: the individual should be placed such that it can perform simple space \& time computations.
\item \textbf{Quality Principle}: the individual must respond to and satisfy various quality factors specified in the environment.
\item \textbf{Diverse Response Principle}: the individual should
not execute its activity along extremely limiting channels.
\item \textbf{Stability Principle}: the individual should never change its
style of action even though the environment may change.
\item \textbf{Adaptability Principle}: the individual must have the ability to change its behaviour mode should the computational price justifies it doing so.
\end{enumerate}


\section{Walk-through of an NIA}
\label{sec:4}

In this section, we will discuss in detail the formulations of a general Swarm Intelligence NIA, and how it is applied in practice to tackle optimisation problems such as route planning. 
 
There are many NIAs which are used for route planning in practice. One of the recent and most effective ones is called \textit{Salp Swarm Algorithm (SSA)}, proposed by Mirjalili et al. in 2016 \cite{survey-16}. It is an efficient algorithm that depicts swarm intelligence by mimicking the behaviour of groups of swarms, which are a sea creature similar to jellyfish. Although it is relatively easy to understand, SSA is an effective algorithm for solving real-world problems. The problems tend to have unknown and challenging search spaces.

\subsection{The inspiration behind Salp Swarm Algorithm}
\label{subsec:4.1}

Salps are a sea creature. They are members of the \textit{Salpidae} \cite{survey-16} family. Salps are known to have a barrel-shaped transparent body similar to jellyfish. Their movement can also be compared to that of jellyfish - they both pump water through their body and move forward as a result of the propulsion.

Salps are known to exhibit \textit{swarming behaviour}, which will be of particular interest to us in this chapter. A group of salps, which is formed during swarming, is known simply as a \textbf{salp chain}. The primary motivation behind this swarming behaviour is unclear, but researchers speculate the motive may be better locomotion of an otherwise vulnerable creature in a hostile environment \cite{survey-17}.

\subsection{Mathematical Model of SSA}
\label{subsec:4.2}

To mathematically model, a natural phenomenon means to devise mathematical equations based on the observations and use these equations in software simulation programs. Before applying SSA, we first divide the salp "population" in two groups: \textit{leaders} and \textit{followers}. The salp at the frontal end of the chain is termed as leader. On the other hand all the other salps that follow the leader are called followers. The job of the leader is to guide the salp chain and the others directly (or indirectly) follow their leader.

We use an $m$-dimensional search space to characterise the locale of salps, where $m$ is the number of variables given. We also assume that a two-dimensional matrix, $x$, is used to store the position of all salps. If the target of the entire chain is defined by a food source, $Food$, then the steps to update the leader's position is given in Eq. (1): 

\begin{equation}
x_j^1 = \begin{cases}
    Food_j + r_1((upper_j - lower_j)r_2 + lower_j), & \text{if $r_3 \geqslant 0$}.\\
    Food_j - r_1((upper_j - lower_j)r_2 + lower_j), & \text{if $r_3 < 0$}.
    \end{cases}
\end{equation}

where the position of the leader in the $j\textsuperscript{th}$ dimension is given as $x_j^1$, the food source is given as $Food_j$, and $upper_j$ and $lower_j$ are the respective upper and lower bounds of the $j\textsuperscript{th}$ dimension. $r_1$, $r_2$, and $r_3$ are three random numbers into the equation.

It can be inferred from Eq. (1) that only the leader updates its position regarding the food source. The most important parameter in SSA is actually $r_1$ as it can be adjusted to balance \textit{exploration} and \textit{exploitation}. $r_1$ can be obtained from Eq. (2):

\begin{equation}
r_1 = 2e^{-(\frac{4n}{N})^2}
\end{equation}

where $N$ is the maximum number of iterations and $n$ is the current iteration.

The other random numbers in Eq. (1), $r_2$ and $r_3$, are uniformly generated in an interval of [0,1]. They define whether the upcoming position in $j\textsuperscript{th}$ dimension would be toward positive direction, or negative direction. They also indicate the step size.

The position of the followers is updated by utilising Newton's laws of motion, Eq. (3):

\begin{equation}
x_j^1 = \frac{1}{2}at^2 + v_0t
\end{equation}

were $x_j^i$ is the position of $i\textsuperscript{th}$ follower salp in $j\textsuperscript{th}$ dimension, for all $i \geqslant 2$. Here, $t$ is time and $v_0$ defines the initial speed. The acceleration, $a = \frac{v_{final}}{v_0}$, and velocity, ${v = \frac{x-x_{0}}{t}}$.

This Eq. (3) is modified relative to the SSA, and can be written is:

\begin{equation}
x_j^i = \frac{1}{2}(x_j^1 + x_j^{i-1})
\end{equation}

where $x_j^i$ shows the position of $i\textsuperscript{th}$ follower salp in $j\textsuperscript{th}$ dimension, for all  $i \geqslant 2$.

The positions of all the salps can be simulated now by using Eq. (1) and Eq. (4).

\subsection{Working of SSA}
\label{subsec:4.3}

To delve into the inner workings of Salp Swarm Algorithm, it is important to realise that the leader salp alone 'leads' the chain and the other salps simply 'follow'. The salps chase the food source. So, if we replace the food source with global optima, then the salp chain automatically moves in that direction. The pseudocode of SSA algorithm can be stated as:

\begin{algorithm}[H]
\caption{SSA Algorithm}
\label{a:ssa}
\begin{algorithmic}[1]
\State Initialise the salp population $x_i$ ($i$ = 1, 2, ..., $m$) considering $upper$ and $lower$ 
		\State \textbf{while} (end condition is not satisfied)
		\State Calculate the fitness of each search agent (salp)
		\State $Food$ = the best search agent
		\State Update $r_1$ by Eq. (2)
		\State \hspace*{0.5cm} \textbf{for} each salp ($x_i$)
		\State \hspace*{1cm} \textbf{if} ($i$ == 1) 
		\State \hspace*{1.5cm} Update the position of the leading salp by Eq. (1)
		\State \hspace*{1cm} \textbf{else}
		\State \hspace*{1.5cm} Update the position of the follower salp by Eq. (4)
		\State \hspace*{1cm} \textbf{end}
		\State \hspace*{0.5cm} \textbf{end}
		\State \hspace*{0.5cm} Amend the salps based on the upper and lower bounds of variables
		\State \textbf{end}
		\State return $Food$
\end{algorithmic}
\end{algorithm}

We start the algorithm with multiple salps in random positions. We then calculate the \textit{fitness} of each salp. The best-fitness salp is assigned to the variable $Food$, and it becomes the food source which is then chased by the salp chain. Now, we update the parameter $r_1$ using Eq. (2). Subsequently, the position of leader salp is updated using Eq. (1). Similarly, the position of the follower salp is updated using Eq. (4), as described in Section \ref{subsec:4.2}. We perform boundary check at every step, wherein we check if any salp goes outside the boundaries of the environment. If it does, it is brought back inside the boundary. We execute all the other steps iteratively, until a suitable satisfaction condition is reached.

One thing to note here is that salp chain can determine a better solution to the problem through exploration and exploitation. Hence, the food source is updated during this process. Thus, the salp chain should be able to chase a moving food source, making it useful in a dynamic target environment.

\subsection{Route Planning using SSA}
\label{subsec:4.4}

\begin{figure}[h]
\centering
\includegraphics[scale=0.20]{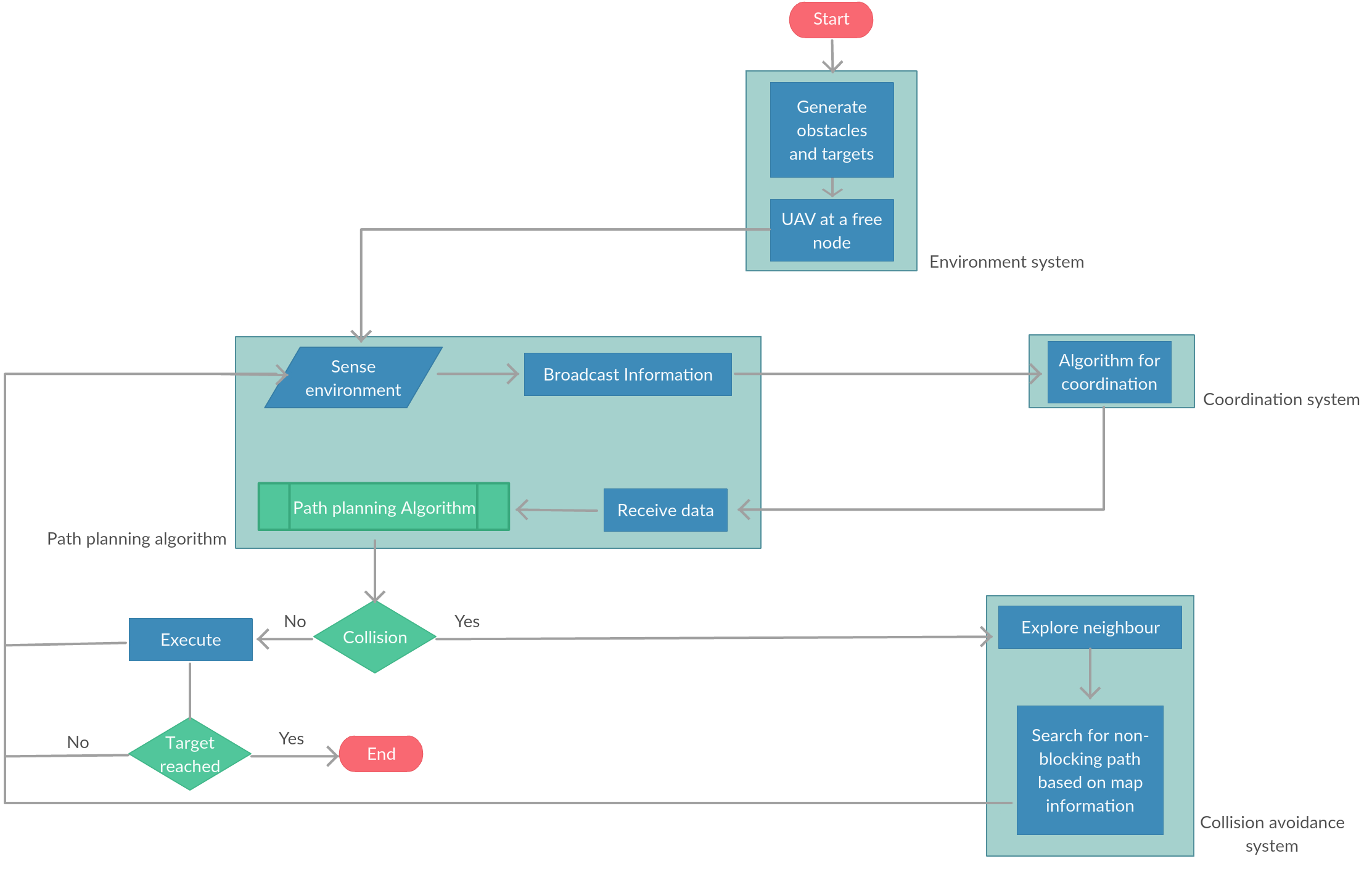}
\caption{Methodology of applying SSA to route planning}
\label{fig:ssa-meth}      
\end{figure}

Figure \ref{fig:ssa-meth} depicts the methodology undertaken when applying SSA to route planning. In the route planning phase, the robot will sense the environment and broadcast this information to the semi-centralised server, which is employed for achieving coordination task among the robots. The received data from the server is used for route planning and moving towards the target. If there is no collision detected, the robot is moved to the required position and checked for target reachability. If the collision is detected collision avoidance algorithm is run, which includes a different strategy for dealing with the static and dynamic obstacles. After collision avoidance, the route planning system is kicked again, and it continues.

\begin{figure}
\centering
\includegraphics[scale=0.27]{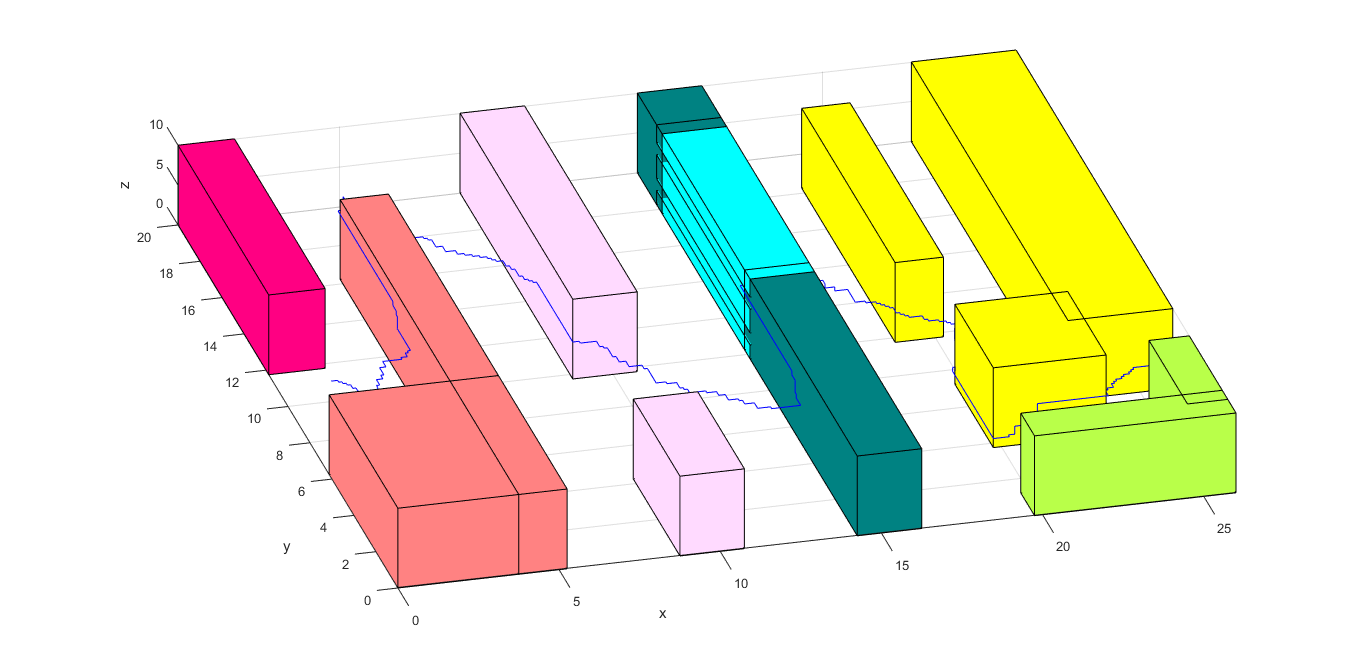}
\caption{Simulation of SSA in 3D Static environment}
\label{fig:ssa-static}      
\end{figure}

The results obtained by applying SSA to 3D static environments is depicted in Figure \ref{fig:ssa-static}. Here, the coloured blocks represent the obstacles of varying shapes and sizes, and the agent moves from left to right, and the blue line traces its route.

\begin{figure}
\centering
\includegraphics[scale=0.57]{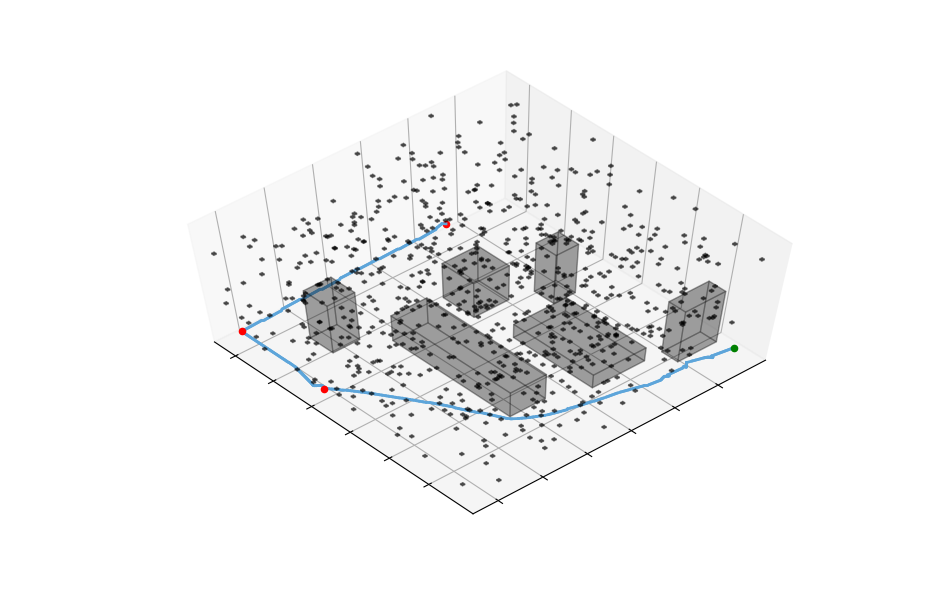}
\caption{Simulation of SSA in 3D Dynamic environment \cite{mds1}}
\label{fig:ssa-dynamic}      
\end{figure}

SSA for dynamic environment is depicted in Figure \ref{fig:ssa-dynamic}. Here, the gray blocks are static obstacles and the black dots contain dynamic obstacles. The red dots represent the targets and the route is traced by the blue line.

\subsection{Comparison of Results}
\label{subsec:4.5}

To compare the results obtained by SSA, several other algorithms were also tested on the same static and dynamic 3D environments: Particle Swarm Optimisation (PSO, described in Section \ref{subsec:5.1}), Glowworm Swarm Optimisation (GSO), and Firefly Algorithm (FA, described in Section \ref{subsec:5.3}). More details about the experimental setup can be found in \cite{survey-pra} and \cite{survey-rkd}.

For the static 3D environment given in Figure \ref{fig:ssa-static}, results obtained are tabulated in Table \ref{tab:1}.

\begin{table}[h]
\centering

\caption{Results in static 3D environment}
\label{tab:1}       
\begin{tabular}{lllll}
\hline\noalign{\smallskip}
\textbf{Algorithm} & \textbf{Population} & \textbf{Iteration} & \textbf{Best Cost} & \textbf{Time}\\
\noalign{\smallskip}\hline\noalign{\smallskip}
PSO & 20 & 25 & \num{8.32e+02} & \num{7.16e+01}\\
GSO & 20 & 25 & \num{9.67e+02} & \num{7.39e+01}\\
SSA & 20 & 25 & \num{7.86e+02} & \num{7.09e+01}\\
\noalign{\smallskip}\hline
\end{tabular}
\end{table}




For the dynamic 3D environment given in Figure \ref{fig:ssa-dynamic}, the results when the above algorithms are applied are tabulated in Table \ref{tab:2}.

\begin{table}[h]
\centering
\caption{Results in dynamic 3D environmentn}
\label{tab:2}       
\begin{tabular}{lllll}
\hline\noalign{\smallskip}
\textbf{Algorithm} & \textbf{Population} & \textbf{Iteration} & \textbf{Best Cost} & \textbf{Time}\\
\noalign{\smallskip}\hline\noalign{\smallskip}
PSO & 20 & 25 & \num{587468} & \num{588}\\
GSO & 20 & 25 & \num{561287} & \num{536}\\
SSA & 20 & 25 & \num{548139} & \num{271}\\
\noalign{\smallskip}\hline
\end{tabular}
\end{table}

As we can see from the above table, the SSA algorithm gives better performance when compared in terms of cost as well as time than all the other algorithms.

%
%

\section{Other Commonly Used NIAs}
\label{sec:5}

Apart from the Salp Swarm Algorithm (SSA), other Swarm Intelligence-based NIAs are also used widely in practice. A few examples include \textit{Particle Swarm Optimisation (PSO)}, \textit{Ant Colony Optimisation (ACO)}, and \textit{Firefly Algorithm (FA)}, among others. This section briefly introduces these three algorithms.

\subsection{Particle Swarm Optimisation}
\label{subsec:5.1}

The PSO technique was proposed by Dr Kennedy and Dr Earhart in 1995. PSO algorithm uses the swarming or social behaviour of flocks of birds and school of fish, as inspiration for particle optimisation \cite{survey-18}. In this technique, each bird or fish is considered as a \textit{particle}, and together the school or the flock acts as a \textit{swarm}. As in the case of SSA, the particles communicate with each other by learning from their experiences and also update themselves by searching the given space and building their memory. PSO would find any route existing in the environment. 

Researchers apply PSO to many optimisation areas. PSO has a unique and straightforward searching mechanism. It is computationally, efficient and relatively easy to implement. Briefly, there are four vectors required to represent a particle in high-dimensional space: the current position, the best position so far, the best position in the entire neighbourhood, and the velocity. The details of PSO are left as further study \cite{survey-15}.

\subsection{Ant Colony Optimisation}
\label{subsec:5.2}

Ant Colony Optimisation (ACO), as the name suggests, aims to depict the behaviour of ants in their natural environment  \cite{survey-1}. Ants form \textit{colonies}, which is analogous to swarm in case of SSA. The ants communicate between each other, and this allows them to compute the shortest route. This corresponds to the route between their nest (hive) \& food source. Ants deposit a chemical substance called \textit{pheromone} along their route. More ants follow the same route, the quantity of pheromone deposited grows and hence other ants can check out the higher pheromone-containing route. Pheromone evaporates with time, so only the most popular routes remain after some time.

As a result, the ACO algorithm works best when the source and destination of the problem are specific and clearly defined. The main steps for the ACO algorithm are as follows:

\begin {enumerate}
\item We generate and initialise the ants.
\item Iterate for each ant (break when we find the goal, or a stopping criteria is met):
\item Accumulate pheromone deposit along the visited route.
\item Daemon activities.
\item Evaporate the pheromone content of less popular route after some time.
\end {enumerate}

\subsection{Firefly Algorithm}
\label{subsec:5.3}

The firefly algorithm was given by Xin-She Yang in 2008 \cite{survey-19}. It finds the global solution of an optimisation problem by on swarm intelligence based on the \textit{flashing behaviour} of fireflies. The firefly's flash operates as a signal to communicate with and warn other fireflies.

Similar to other algorithms of its kind, FA begins by generating random initial populations that can have feasible candidate solutions. The aim is that knowledge is collectively shared among all fireflies in the population. This guides the search for the best solution in the search space. Each particle has the capability to move in a multi-dimensional space and an attractiveness that is actively updated based on the knowledge of other fireflies and neighbours \cite{survey-20}.

%
%

\section{The motivation from nature}
\label{sec:6}

Now that we have given insights into some of the standard Nature-Inspired Algorithms, it is the right time to discuss the rationale behind using them for optimisation problems. NIAs are often more complex and time-taking than conventional algorithms. Moreover, most NIAs are often less intuitive or require a great deal of research before devising the mathematical model to use them correctly. 

However, NIAs are being used rapidly in many optimisation problems. We as humans love to explore and exploit the strengths of nature in almost every domain. A similar act in this domain can help researchers to make the best out of NIA and identify potential solutions for complex real-world scenarios. Moreover, to discover new methods and to improve the previous, the potential merits of interdisciplinary research are highlighted.  

In the following sections, we dive a little deeper into the process of evolution in nature and related consider processes. Further, we describe the importance of metaphor in algorithm research along with, the way nature attains and provides encouragement for creativity. This section focuses on the aspects that we recognise as interesting and relevant, but it should not be considered as the only exhaustive resource which nature provides us.

\subsection{Natural Selection and Optimisation}
\label{subsec:6.1}

Darwin's theory is quite often quoted as, one of, if not the most important scientific theory in the 19\textsuperscript{th} century. This can be attributed to the fact that this theory uses relatively simple rules of \textit{reproduction}, \textit{mutation} and \textit{selection}, and combines then in a powerful way.\\

According to this theory, the fittest animals are ones who have superior problem-solving abilities. 
These abilities are the source of \textit{strong} inspiration for many popular meta-heuristic based optimisation algorithms. Although it is hard to say precisely what is being optimised by natural selection, we do observe it has produced certain useful features - \textit{adaptation}, \textit{efficiency} and \textit{generality}. A detailed analysis of these features is left as further study \cite{survey-21}.\\

\subsection{Complex Systems and Emergent Behaviour}
\label{subsec:6.2}

The presence of high-level function that develops due to the interaction among elements is termed as emergent behaviour. If the behaviour of the whole and not merely of its parts \textit{(i.e., no-linear aggregate behaviour)}, then it is said to be a case of emergent behaviour. This, however, does not rule out a deterministic relationship between the parts and the whole.

Nature has the advantage of trial and error - a few hundred million years as such - and it enables nature for producing useful emergent behaviour by devising some rules. The rules often can be a student by an algorithm designer - with the possible help of a biological researcher - and be extracted into mathematical models, useful for generating computer algorithms. 

In this way, algorithms that are simple enough to code can be generated from relatively complex processes. These emergent approaches to optimisation are also well suited to parallel computation, and as such, capable of utilising recent developments in the field.

\subsection{Natural Metaphors}
\label{subsec:6.3}

We, humans, tend to search for patterns in every facade of life. Researchers have found similar patterns between domain-specific problems and those which exist in nature. This has made us realise how nature solves a particular problem and harness that knowledge. Additionally, this \textit{natural metaphor} enables us to think abstractly about different solutions. 

%
%

\section{Summary}
\label{sec:7}

It may come as no surprise that NIAs have a wide variety of applications in real-life scenarios. There has been tremendous growth in the domain of nature-inspired science in recent decades. These algorithms are often used for investigating nature by using computer simulations. This chapter provided a general walk-through of a nature-inspired algorithm as well as give their detailed analysis and implementation on route planning problems. We hope that more researchers would take up this topic in the future, and the development of new and upgraded algorithms to solve fundamental human problems would be more commonplace.

In conclusion, we would like to quote David Perkins in \textit{Archimedes' Bathtub} by saying \cite{survey-22}: \begin{quotation}Mother Nature may re-purpose, but do we see the full pattern of breakthrough thinking in nature -- the long search, little apparent progress, the precipitating event, some non-mental equivalent of the cognitive snap, and transformation? Arguably, yes! \end{quotation}

\end{document}